\documentclass[10pt,twocolumn,letterpaper]{article}

\usepackage[pagenumbers]{cvpr} 

\definecolor{cvprblue}{rgb}{0.21,0.49,0.74}
\usepackage[pagebackref,breaklinks,colorlinks,allcolors=cvprblue]{hyperref}

\def\paperID{} 
\def\confName{}
\def\confYear{}

\title{LAB-Det: Language as a Domain-Invariant Bridge for Training-Free One-Shot Domain Generalization in Object Detection}

\author{
{Xu Zhang\textsuperscript{1} \quad Zhe Chen\textsuperscript{2} \quad Jing Zhang\textsuperscript{3} \quad Dacheng Tao\textsuperscript{1}} \\
\normalsize \textsuperscript{1}The University of Sydney, Australia \quad 
\normalsize \textsuperscript{2}La Trobe University, Australia \quad 
\normalsize \textsuperscript{3}Wuhan University, China \\ 
\tt\small 
xzha0930@uni.sydney.edu.au \quad zhe.chen@latrobe.edu.au \\ 
\tt\small jingzhang.cv@gmail.com \quad dacheng.tao@gmail.com
}

\begin{document}
\maketitle

\begin{abstract}
Foundation object detectors such as GLIP and Grounding DINO excel on general-domain data but often degrade in specialized and data-scarce settings like underwater imagery or industrial defects. Typical cross-domain few-shot approaches rely on fine-tuning scarce target data, incurring cost and overfitting risks. We instead ask: Can a frozen detector adapt with only one exemplar per class without training? To answer this, we introduce training-free one-shot domain generalization for object detection, where detectors must adapt to specialized domains with only one annotated exemplar per class and no weight updates. To tackle this task, we propose LAB-Det, which exploits Language As a domain-invariant Bridge.  
Instead of adapting visual features, we project each exemplar into a descriptive text that conditions and guides a frozen detector. This linguistic conditioning replaces gradient-based adaptation, enabling robust generalization in data-scarce domains.
We evaluate on UODD (underwater) and NEU-DET (industrial defects), two widely adopted benchmarks for data-scarce detection, where object boundaries are often ambiguous, and LAB-Det achieves up to 5.4 mAP improvement over state-of-the-art fine-tuned baselines without updating a single parameter. These results establish linguistic adaptation as an efficient and interpretable alternative to fine-tuning in specialized detection settings.
\end{abstract}    
\section{Introduction}
\label{sec:intro}

\begin{figure}[t]
  \centering
   \includegraphics[width=1.0\linewidth]{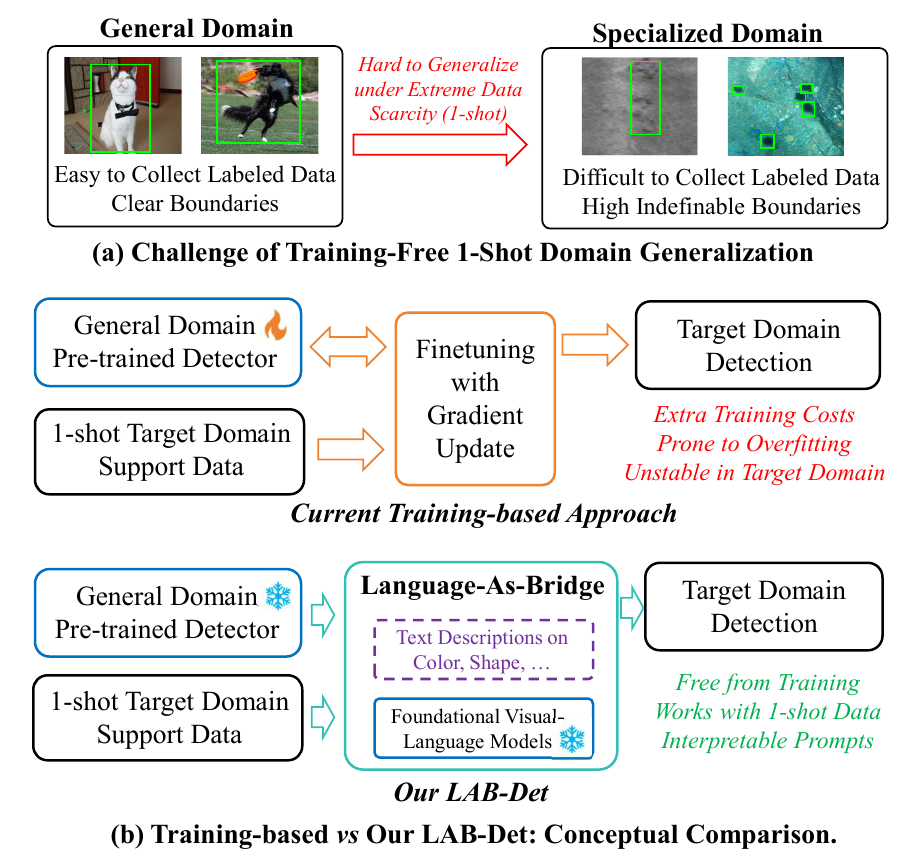}
   \caption{
   Challenge and conceptual comparison of training-free 1-shot domain generalization for object detection. (a) In specialized domains (e.g., underwater or industrial), data can be very scarce (e.g., only one labeled exemplar per class) and object boundaries are ambiguous, making general-domain detectors fail under severe domain gaps. (b) Training-based approaches rely on fine-tuning with gradient updates, which incur extra cost, overfit under 1-shot supervision, and remain unstable in high-IB domains. In contrast, LAB-Det leverages language as a domain-invariant bridge: support exemplars are re-expressed as descriptive text and injected into frozen vision–language models, enabling training-free, interpretable, and robust target-domain detection.
   }
   \label{fig:title}
\end{figure}

Object detection has evolved significantly, transitioning from convolutional neural networks (CNNs) to end-to-end transformer-based architectures \citep{fast_rcnn, detr}. With these advances, detection has become a cornerstone in vision applications, powering industrial inspection, environmental monitoring, medical imaging, and autonomous systems. Yet, detectors trained on large curated sources often fail in specialized domains such as underwater imagery or steel surface inspection, where visual styles diverge sharply and labeled data are scarce. In these settings, detection is not optional but a necessary function. For example, detectors can help identify structural defects in pipelines to prevent costly failures or monitor marine species for ecological sustainability. However, collecting and annotating data in such specialized domains is expensive: industrial defects require expert metallurgists, underwater images demand costly acquisition, and in both cases instances are rare and boundaries are ambiguous \citep{uodd, neu_det}. These realities make large-scale annotation infeasible, underscoring the need for adaptation methods that can work with minimal supervision.

Existing paradigms only partially address this challenge. Few-shot object detection (FSOD) adapts to novel categories with minimal labeled examples \citep{meta_rcnn, fsrw}, but assumes support and query images share the same domain, leaving it brittle under large visual shifts. Cross-domain detection (CDOD) \citep{da_frcnn, progressive_da} tackles domain gaps such as weather or modality changes, but typically relies on abundant unlabeled target data, which could be unrealistic in specialized domains. Cross-domain few-shot detection (CD-FSOD) \citep{cd_vito} combines these challenges. While effective to some extent, CD-FSOD methods still depend on fine-tuning, which is computationally costly, hyperparameter-sensitive, and unstable under extreme data scarcity. In fact, as we show in experiments, strong CD-FSOD baselines collapse on UODD \citep{uodd} and NEU-DET \citep{neu_det}, underscoring the need for a fundamentally different solution. As our experiments confirm, even strong CD-FSOD baselines collapse on UODD \citep{uodd} and NEU-DET \citep{neu_det}, highlighting the need for a fundamentally different approach.
Meanwhile, foundation models such as GLIP \citep{glip}, MDETR \citep{mdetr}, and Grounding DINO \citep{grounding_dino} have become widely adopted for their strong general-domain performance. Yet, these models are pre-trained on datasets that exclude specialized environments, and fine-tuning them on one or two target examples is ineffective \citep{cd_vito}. Thus, despite their potential, foundation detectors remain difficult to deploy in precisely the domains where data is most expensive to obtain.

A promising direction comes from vision-language research: models like CLIP \citep{clip} and ALIGN \citep{align} show that text can connect visual features and generalize robustly across domains. Building on this idea, we hypothesize that language can serve as a domain-invariant bridge for detection. By re-describing exemplars in language, low-level style variations such as color casts or texture noise are abstracted away, consistent with theoretical analyses of domain divergence \citep{ben2010theory}. This motivates a brand new paradigm: rather than adapting visual features through fine-tuning, we adapt foundation detectors by projecting target exemplars into text and using those descriptions to guide detection in a training-free manner.

To this end, we introduce \emph{training-free one-shot domain generalization} for object detection, where each category provides a single annotated exemplar, but NO weight updates are permitted. We further propose \textbf{LAB-Det} (Language As a domain-invariant Bridge for Detection), which projects each exemplar into descriptive phrases using the Describe Anything Model (DAM) \citep{dam}. These domain-derived prompts condition a frozen Grounding DINO \citep{grounding_dino} to generate proposals and assign categories, while a product-of-experts fusion with BLIP \citep{blip} selectively stabilizes small or ambiguous detections. By leveraging language to connect exemplars and frozen detectors, LAB-Det enables robust domain generalization in a fully training-free manner without any gradient updates.

In summary, our main contributions are:
\begin{itemize}
\item We formulate training-free one-shot domain generalization for object detection, a new setting that reflects more realistic deployment constraints in specialized domains, where only one labeled exemplar per class is available and no parameter updates are considered.

\item We present LAB-Det, a framework that leverages Language as a Domain-invariant Bridge. By projecting each exemplar into descriptive text with the Describe Anything Model and conditioning a frozen Grounding DINO, LAB-Det achieves robust generalization. A product-of-experts fusion with BLIP further enhances reliability for small or ambiguous objects.

\item Through extensive experiments on UODD (underwater) and NEU-DET (industrial defects) (\emph{i.e.,} widely adopted data-scarce benchmarks), we show that LAB-Det surpasses state-of-the-art fine-tuned CD-FSOD baselines \citep{cd_vito}, demonstrating that linguistic adaptation can outperform gradient-based fine-tuning in extremely low-data domains.
\end{itemize}
\section{Related Work}
\subsection{Few-Shot and Cross-Domain Object Detection}

Few-shot object detection (FSOD) enables detectors to recognize novel categories with limited labeled examples. Meta-learning approaches, such as Meta R-CNN \citep{meta_rcnn} and FSRW \citep{fsrw}, learn transferable meta-knowledge for fast category adaptation. Transfer-based methods like TFA \citep{tfa}, FSCE \citep{fsce}, and DeFRCN \citep{defrcn} instead fine-tune pre-trained detectors, achieving strong in-domain performance. However, FSOD generally assumes source and target share a similar visual style, and its effectiveness drops sharply under large domain shifts.  
Cross-domain object detection (CDOD) explicitly addresses style shifts, such as changes in weather or imaging modality, typically via unsupervised domain adaptation. DA Faster R-CNN \citep{da_frcnn} aligns features through a domain classifier, while Progressive DA \citep{progressive_da} adapts on synthetic or pseudo-labeled target data. These methods, however, require abundant unlabeled target images, which is impractical in specialized domains where data collection is costly and instances are rare.  
Cross-domain few-shot object detection (CD-FSOD) integrates both challenges: minimal supervision and significant domain gaps. CD-ViTO \citep{cd_vito} enhances DE-ViT \citep{devit} with learnable features and domain prompting to mitigate degradation on benchmarks like UODD and NEU-DET. Other approaches, including Distill-cdfsod \citep{distill_cdfsod} and MoFSOD \citep{mofsod}, also rely on fine-tuning, while AcroFOD \citep{acrofod} and AsyFOD \citep{asyfod} assume overlapping categories across domains, limiting applicability to novel-class settings.  

Overall, existing CD-FSOD methods remain bound to gradient-based updates, which are computationally costly, hyperparameter-sensitive, and brittle when only a single target exemplar is available. This motivates the need for a training-free alternative that can bridge severe domain gaps under extreme data scarcity.

\subsection{Vision-Language Object Detection}

Vision-language models (VLMs) and foundation detection models leverage text to enhance generalization, motivating our prompt-based framework. VLMs like CLIP \citep{clip}, ALIGN \citep{align}, and BLIP \citep{blip} demonstrate that text connects visual features across domains, enabling robust generalization in classification tasks. Building on this, open-vocabulary detection methods utilize VLM capabilities to connect visual features and categories. PromptDet \citep{promptdet}, DetPro \citep{detpro}, and RegionCLIP \citep{regionclip} exploit CLIP’s text-visual alignment to achieve open-vocabulary classification, requiring substantial training data to align box regions with categories. Detic \citep{detic} incorporates additional image-level data for co-training. In contrast, foundation detection models like GLIP \citep{glip}, MDETR \citep{mdetr}, and Grounding DINO \citep{grounding_dino} use vision-language pre-training to align box regions with text prompts, achieving strong zero-shot performance on general-domain tasks, earning widespread community adoption. However, their pre-training datasets, primarily general-domain corpora, limit their effectiveness in specialized domains like underwater or industrial scenes \citep{cd_vito, uodd, neu_det}, where data collection and annotation are challenging due to environmental constraints and the need for expert labeling. Training such models is resource-intensive, underscoring the importance of researching adaptation strategies for these foundation models in data-scarce, specialized domains. In contrast, our LAB-Det leverages a single support image’s descriptive prompts to guide frozen foundation models, offering a data-efficient, training-free solution for data-scarce, specialized domains \citep{cd_vito}.

\section{Method}
\subsection{Definition of Training-Free One-Shot Domain Generalization}

We consider the problem of adapting a pre-trained detector to a specialized target domain where annotation is prohibitively costly. Formally, let $\mathcal{X}$ denote target-domain images with category set $\mathcal{Y} = \{1, \dots, C\}$. For each class $c \in \mathcal{Y}$, we are given a single labeled exemplar in the form of a support triple $(x_c, b_c, t_c)$, where $x_c$ is an image, $b_c = (x_1, y_1, x_2, y_2)$ a bounding box, and $t_c$ the class name. The support set is $\mathcal{S} = \{(x_c, b_c, t_c)\}_{c=1}^C$. Importantly, no further target labels are available, reflecting the real cost of annotation in domains such as underwater monitoring or industrial inspection.  
In this setting, the detector $f_{\theta_0}$ is pre-trained on generic source data and must remain \emph{frozen}: no gradient propagation or weight updates are allowed. Thus, adaptation is possible only through training-free mechanisms that leverage the one-shot supports for conditioning. At test time, given an unseen target image $x \in \mathcal{X}$, the system must output detections $(b_i, \hat{c}_i, s_i)$ for classes in $\mathcal{Y}$. Evaluation follows COCO \citep{coco} mAP on the target test set $\mathcal{T}$.  

We borrow the idea of the established CD-ViTO protocol \citep{cd_vito} and use two widely adopted data-scarce benchmarks: UODD \citep{uodd} for underwater species and NEU-DET \citep{neu_det} for industrial defects. Each provides one annotated exemplar per class and official test splits, enabling fair comparison with fine-tuned CD-FSOD baselines. Note that we do not consider finetuning as used in CD-ViTO. 

\subsection{LAB-Det Framework Overview}

\begin{figure*}[t]
  \centering
   \includegraphics[width=1.0\linewidth]{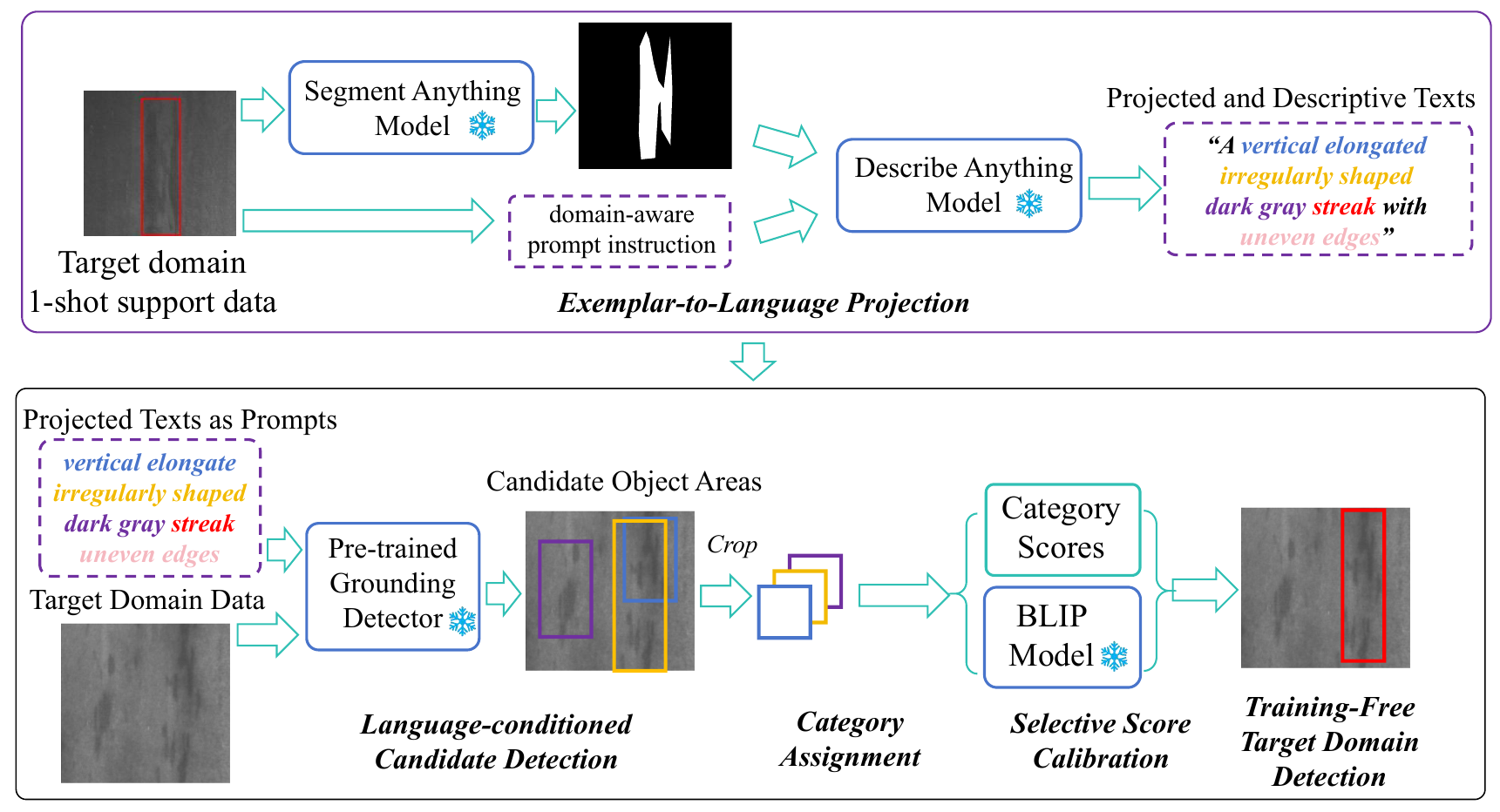}
   \caption{
   Overview of LAB-Det. Top: exemplar-to-language projection. A single annotated exemplar is segmented by SAM and described by DAM under a domain-aware prompt, yielding natural-language phrases (e.g., “rough texture”). Bottom: these phrases condition a frozen detector (e.g., Grounding DINO) to generate candidate boxes and phrase scores. Category scores are obtained by averaging phrase scores, and an optional BLIP-based calibration refines small or ambiguous detections. The entire pipeline is training-free and interpretable.
   }
   \label{fig:pipeline}
\end{figure*}

To address this task, we propose \textbf{LAB-Det} (\emph{Language as a Domain-invariant Bridge for Detection}), a fully training-free framework that re-frames adaptation as linguistic conditioning of a frozen detector. Rather than updating parameters, LAB-Det projects each one-shot exemplar into descriptive language and uses these prompts to guide detection.  
As illustrated in \cref{fig:pipeline}, LAB-Det consists of four stages: (1) \emph{Exemplar-to-Language Projection}, which converts each support box into descriptive phrases using the Describe Anything Model (DAM) \citep{dam}; (2) \emph{Language-conditioned Candidate Detection}, where a frozen Grounding DINO \citep{grounding_dino} uses these phrases to propose candidate boxes; (3) \emph{Category Assignment}, which aggregates phrase-level scores into class predictions; and (4) \emph{Selective Score Calibration}, which fuses predictions with BLIP \citep{blip} alignment for small or ambiguous objects.  

This pipeline leverages language as a domain-invariant bridge, enabling robust generalization in specialized domains without any parameter updates, while maintaining interpretability at the phrase level.

\subsubsection{Exemplar-to-Language Projection}

For each category $c \in \{1,\dots,C\}$ in the target domain, we are given a single support triple $(x_c, b_c, t_c)$: $x_c$ is the support image, $b_c = (x_1,y_1,x_2,y_2)$ is the annotated bounding box of the exemplar, $t_c$ is the category name (e.g., \emph{``scallop''} in UODD or \emph{``scratches''} in NEU-DET).  
The domain $d \in \{\text{underwater}, \text{industrial defect}\}$ provides contextual information (e.g., whether the scene is an underwater habitat or a steel surface).  

We first extract the tight region of the target object with the help of SAM \citep{sam}, with $b_c$ as the input prompt. Then, the related cropped patch, together with a domain-aware instruction,  

\texttt{"This is a \textbf{$d$} image. The masked object is a \textbf{$t_c$}. Describe it in one short sentence using the word \textbf{$t_c$}"},  

\noindent are provided to the frozen Describe Anything Model (DAM) \citep{dam}. DAM then outputs a descriptive sentence $p_c$, such as:

\texttt{"A vertical, elongated, irregularly shaped dark gray streak with a rough texture and uneven edges."}. 

\noindent The generated description $p_c$ can be naturally utilized as a set of phrases $\{p_{c,m}\}_{m=1}^{M_c}$. This method capitalizes on Grounding DINO's inherent sensitivity to phrasal units, bypassing the need for dedicated parsing. The model's output is often inherently composed of semantically meaningful spans such as noun or adjective phrases (e.g., ``elongated irregularly shape,'' ``rough texture,'' ``dark gray''), which naturally arise from grouping adjacent modifiers with their heads and serve as the effective input for the detector.
This results in a variable-length phrase set for each class, where $M_c$ denotes the number of phrases.  
These multi-word descriptors allow Grounding DINO to exploit fine-grained semantic cues beyond the raw class name.

\subsubsection{Language-conditioned Candidate Detection}

Given the phrase library, we prompt the frozen detector $f_{\theta_0}$ (Grounding DINO) with: 
\begin{equation}
\mathcal{P} = \{ p_{c,m} \mid c = 1,\dots,C, \, m = 1,\dots,M_c \}.
\end{equation}
Taking $\mathcal{P}$ as input prompt for $f_{\theta_0}$, on an unseen target image $x$, the detector could return $N$ candidate boxes $\{b_i\}_{i=1}^N$ along with phrase-level relevance scores $s^{\text{gd}}_{i,c,m} \in [0,1]$, each measuring how well candidate $b_i$ matches each phrase $p_{c,m}$. 
Collecting all scores yields  
\begin{equation}
\mathbf{S}^{\text{gd}} = \{ s^{\text{gd}}_{i,c,m} \mid i=1\dots N,\, c=1\dots C,\, m=1\dots M_c \},
\end{equation}
which we refer to as the phrase–box–category score tensor. Remind that $N$ is the number of candidate boxes per image, $C$ is the number of categories, and $M_c$ is the number of phrases (tokens) extracted for category $c$. Then, after the initial detection on an image, the score $s^{\text{gd}}_{i,c,m}$ actually quantifies the degree to which a candidate box $b_i$ corresponds to a phrase of $p_{c,m}$. The tensor $\mathbf{S}^{\text{gd}}$ therefore represents the complete collection of phrase-level grounding scores across all candidate boxes, categories, and phrases.  

We propose that phrase granularity improves robustness under domain shifts: if color descriptors fail under underwater lighting, other attributes such as shape or texture (e.g., “elongated” or “rough”) can still activate correct detections. Importantly, all detector weights $\theta_0$ remain frozen; adaptation arises purely from linguistic prompts.

\subsubsection{Category Assignment}

To consolidate phrase-level evidence, we compute a category score for each candidate box $b_i$ as the average of all phrase scores associated with category $c$:  
\begin{equation}
O_{i,c} = \frac{1}{M_c} \sum_{m=1}^{M_c} s^{\text{gd}}_{i,c,m}.
\end{equation}
We propose that averaging is more stable than a hard maximum: if one attribute phrase fails, others (e.g., shape or texture) can still contribute to the score.  

We then flatten $O \in \mathbb{R}^{N \times C}$ and select the global top-$K$ highest-scoring box–category pairs. Each retained detection is represented as $(b_i, \hat{c}_i, s_i)$, where $s_i = O_{i,\hat{c}_i}$ is the objectness score, $\hat{c}_i$ is the predicted label, and $b_i$ the box coordinates. This procedure is fully parameter-free and interpretable: every predicted label can be traced back to the supporting phrases.

\subsubsection{Selective Score Calibration}

Often, small objects (e.g., $\operatorname{area}(b_i) < 32^2$ pixels following the COCO definition) yield unstable scores due to limited visual evidence. To mitigate this, we apply a selective calibration using BLIP \citep{blip}.  

For each retained detection, represented as $(b_i, \hat{c}_i, O_{i,\hat{c}_i})$, where $b_i$ is the detected box, $\hat{c}_i$ as its assigned category label (selected in the category-assignment stage), and $O_{i,\hat{c}_i}$ as the corresponding confidence score. We then query BLIP with the cropped region $b_i$ and the full category description $p_{\hat{c}_i}$, obtaining an alignment score $A_i \in [0,1]$. The final confidence is computed as  
\begin{equation}
S_i = (1-\lambda) O_{i,\hat{c}_i} + \lambda A_i .
\label{eq:si}
\end{equation}
Here $\lambda=0.02$ is deliberately set small to account for scale differences: BLIP’s alignment scores can be relatively uncalibrated compared to detector objectness, and over-weighting them could suppress true positives. A small $\lambda$ ensures that BLIP serves as a lightweight corrective signal without overwhelming the detector’s own confidence, stabilizing noisy phrase matches on small or ambiguous objects. 
This fusion strategy gives higher confidence only to boxes that are simultaneously supported by the detector’s objectness score and BLIP’s alignment score, thereby suppressing noisy detections on small or ambiguous objects while leaving larger boxes unaffected. Because BLIP is used in a frozen form, the procedure remains fully training-free. A more formal view of this mechanism as a product-of-experts fusion is provided in the next section (Sec.~\ref{Sec:dis}).

\subsection{Why Language Helps: Theoretical Intuition}
\label{Sec:dis}

LAB-Det re-frames cross-domain adaptation as linguistic conditioning of a frozen detector. Instead of adapting parameters with gradient updates, a single exemplar is projected into language; phrase-grounded proposals are generated; category scores are aggregated; and a lightweight vision–language check calibrates small objects. Despite involving no parameter updates, the method generalizes across domains with severe visual shifts. To understand why this works, we provide two complementary views.  

\paragraph{Product-of-Experts View.}  
Consider a detection box $b_i$ assigned to category $\hat{c}_i$ with detector confidence $O_{i,\hat{c}_i}$ (from Grounding DINO) and alignment score $A_i$ (from BLIP). We can define the corresponding energies as:
\begin{equation}
E^{\text{obj}}_{i,\hat{c}_i} = -\log O_{i,\hat{c}_i}, \quad 
E^{\text{vl}}_{i,\hat{c}_i} = -\log A_i .
\end{equation}
Based on Eq. \ref{eq:si}, the final confidence is
score $S_i$, so the related energy is then:
\begin{equation}
S_i \propto \exp\!\left(-E^{\text{obj}}_{i,\hat{c}_i} - E^{\text{vl}}_{i,\hat{c}_i}\right).
\end{equation}
This corresponds to the classic \emph{product-of-experts} (PoE) formulation \citep{hinton2002poe}: a prediction receives high confidence only when endorsed simultaneously by both the objectness expert (Grounding DINO) and the vision–language expert (BLIP).  
This is distinct from a standard ensemble (e.g., averaging scores), which rewards agreement but does not strictly penalize disagreement. PoE is stricter: if either expert is uncertain, the combined energy remains high, suppressing false positives. Crucially, LAB-Det achieves this PoE-style fusion without training (e.g., simply by weighted addition in log space), which provides a transparent, training-free mechanism that stabilizes small or ambiguous detections.  

\paragraph{Domain-Adaptation Bound View.}  
From another perspective, let $\varepsilon_T(h)$ denote the risk of hypothesis $h$ on the target distribution $P_T$, and $\varepsilon_S(h)$ the source risk. The Ben–David bound \citep{ben2010theory} states:
\[
\varepsilon_T(h) \le \varepsilon_S(h) + \tfrac{1}{2} d_{\mathcal{H}\Delta\mathcal{H}}(P_S,P_T) + \lambda ,
\]
where $d_{\mathcal{H}\Delta\mathcal{H}}$ measures distributional divergence between source and target, and $\lambda$ is the joint error of the optimal hypothesis across both domains.  
In our setting, projecting exemplars into language can reduce $d_{\mathcal{H}\Delta\mathcal{H}}$: linguistic descriptions are more invariant to pixel-level distortions (e.g., underwater color casts, industrial textures) than raw visual features, narrowing the effective gap between source and target. The selective fusion further lowers $\lambda$ by requiring agreement between complementary experts, thereby reducing the chance of systematic errors. Together, these mechanisms explain why LAB-Det can achieve lower target-domain risk $\varepsilon_T(h)$ without modifying detector parameters.  

\paragraph{Summary.}  
The above analysis clarifies why language helps: projecting exemplars into text reduces domain divergence, and combining frozen experts in a PoE manner lowers joint errors. LAB-Det thus provides an interpretable, theoretically grounded, and fully training-free solution to one-shot domain generalization in detection.

\section{Experiments}
\subsection{Experimental Setup}
\subsubsection{Datasets and Evaluation Metrics}

We evaluate LAB-Det on two specialized domains highlighted by CD-ViTO \citep{cd_vito} as having the highest indefinable boundary (IB) scores:  \textbf{UODD} \citep{uodd}, an underwater dataset with three categories (\emph{sea urchins, sea cucumbers, scallops}) and 506 images containing 3,218 boxes. \textbf{NEU-DET} \citep{neu_det}, an industrial surface defect dataset with six defect types (\emph{rolled-in scale, crazing, pitted surface, patches, inclusion, scratches}) across 360 images with 834 boxes.  

Both datasets are small, costly to annotate, and contain blurred boundaries and domain-specific artifacts that diverge strongly from web-scale training data. Following CD-ViTO’s protocol, we adopt their official one-shot support and test splits: one annotated exemplar per class is provided as support, and all official test images are used for evaluation. No additional data is used.  
Performance is reported using COCO mean Average Precision (mAP) \citep{coco}, averaged over IoU thresholds from 0.50 to 0.95.

\subsubsection{Implementation Details}
Since DAM \citep{dam} does not directly accept bounding boxes, we first convert each support box $b_c$ into a mask using SAM-ViT-Huge \citep{sam}. The mask and a domain-aware instruction (\emph{``This is a \textless domain\textgreater\ image. The masked object is a \textless category\textgreater. Describe it in one short sentence using the word \textless category\textgreater.''}) are fed to DAM-3B, which outputs a single descriptive sentence containing multiple phrases. This step is performed offline, and only the resulting text is stored.  

For detection, we use the official Grounding DINO checkpoint \citep{grounding_dino} with Swin-T \citep{liu2021swin} as the image encoder, BERT \citep{devlin2018bert} as the text encoder, and DINO \citep{zhang2022dino} as the detection head. All parameters are frozen. Each target image is processed with the full phrase set $\mathcal{P}$, and we retain the top $K=300$ scored proposals.  
For small proposals ($< 32^2$ pixels), we query the BLIP model \citep{blip} with the cropped box and the corresponding description $p_{\hat{c}}$, producing an alignment score $A_i \in [0,1]$. Final confidence is computed via \cref{eq:si}.  

All components (Grounding DINO, SAM, DAM, BLIP) are used with publicly released checkpoints and default hyperparameters. Crucially, no gradient propagation or parameter updates are performed, ensuring that LAB-Det fully adheres to the training-free paradigm.

\subsection{Results and Analysis}

To contextualize LAB-Det within existing approaches, we compare against four representative families of detectors, following the taxonomy in Distill-cdfsod \citep{distill_cdfsod} and CD-ViTO \citep{cd_vito}:  
\textbf{(1) Typical FSOD.} We include four standard few-shot detectors, e.g., Meta-RCNN \citep{meta_rcnn}, TFA \citep{tfa}, FSCE \citep{fsce}, and DeFRCN \citep{defrcn}, to assess how conventional fine-tuning methods perform when domain gaps are extreme.  
\textbf{(2) ViT-based OD.} To examine the effect of large-scale vision transformers, we evaluate ViTDeT \citep{vitdet} fine-tuned on the one-shot support set (ViTDeT-FT), following the CD-ViTO adaptation protocol.  
\textbf{(3) Open-vocabulary FSOD/OD.} We consider two open-set baselines: DE-ViT \citep{devit} and Detic \citep{detic}, each in both frozen and fine-tuned forms (DE-ViT/DE-ViT-FT and Detic/Detic-FT). These test whether pre-trained open-vocabulary systems can bridge domain gaps without additional supervision.  
\textbf{(4) CD-FSOD.} Finally, we include two dedicated cross-domain few-shot methods: Distill-cdfsod \citep{distill_cdfsod} and CD-ViTO \citep{cd_vito}, which directly tackle both the domain shift and limited supervision.  

For fairness, baseline scores are taken from the original papers or from the reproducible results released by Distill-cdfsod and CD-ViTO, both of which follow a unified one-shot protocol with public training settings. This ensures all methods, including LAB-Det, are evaluated on the same support/test splits and under comparable conditions.

\begin{table}[t]
\centering
\caption{\textbf{UODD} (one-shot) comparison. Training: $\checkmark$ represents fine-tuned, $\times$ represents training-free. The best results are highlighted in \textbf{bold}.}
\begin{tabular}{lcc}
\toprule
Method & Training & mAP (\%) \\
\midrule
Meta-RCNN \citep{meta_rcnn}        & $\checkmark$ & 3.6 \\
TFA w/ cos \citep{tfa}             & $\checkmark$ & 4.4 \\
FSCE \citep{fsce}                  & $\checkmark$ & 3.9 \\
DeFRCN \citep{defrcn}              & $\checkmark$ & 4.5 \\
\midrule
ViTDeT-FT \citep{vitdet}           & $\checkmark$ & 4.0 \\
Detic \citep{detic}                & $\times$      & 0.0 \\
Detic-FT \citep{detic}             & $\checkmark$ & 4.2 \\
DE-ViT \citep{devit}               & $\times$      & 1.5 \\
DE-ViT-FT \citep{devit}            & $\checkmark$ & 2.4 \\
\midrule
Distill-cdfsod \citep{distill_cdfsod} & $\checkmark$ & 5.9 \\
CD-ViTO \citep{cd_vito}            & $\checkmark$ & 3.1 \\
\midrule
\textbf{LAB-Det (ours)}          & $\times$      & \textbf{6.6} \\
\bottomrule
\end{tabular}
\label{tab:uodd}
\end{table}

\subsubsection{Experiments on UODD}

On the UODD dataset, which features underwater scenes with severe domain shifts and blurred boundaries, LAB-Det achieves 6.6 mAP in the one-shot setting, outperforming all baselines (\cref{tab:uodd}). 
Compared with FSOD baselines, fine-tuned approaches such as DeFRCN \citep{defrcn} reach only 4.5 mAP, reflecting their difficulty in bridging large visual gaps with minimal supervision.  Compared with open-vocabulary methods, training-free variants easily collapse (e.g., Detic: 0.0 mAP; DE-ViT: 1.5 mAP) due to pre-training mismatch, while fine-tuned versions improve slightly but are still far behind (e.g., Detic-FT: 4.2 mAP, DE-ViT-FT: 2.4 mAP). Compared with CD-FSOD baselines, even specialized methods fall short: LAB-Det surpasses Distill-cdfsod \citep{distill_cdfsod} by +0.7 mAP and CD-ViTO \citep{cd_vito} by +3.5 mAP.  

Overall, LAB-Det’s advantage is striking: it outperforms every fine-tuned baseline while requiring no parameter updates, unlike Meta-RCNN \citep{meta_rcnn}, TFA \citep{tfa}, and ViTDeT-FT \citep{vitdet}, which rely on gradient descent over the support image. We attribute this gain to the domain-aware prompts generated from exemplars, which inject shape and texture cues better suited to underwater distortions than raw class names. These results affirm our central claim: language serves as a domain-invariant bridge, enabling training-free generalization that can outperform fine-tuning in data-scarce, high-shift environments.

\begin{table}[t]
\centering
\caption{\textbf{NEU-DET} (one-shot) comparison. Training: $\checkmark$ represents fine-tuned, $\times$ represents training-free. The best results are highlighted in \textbf{bold}.}
\begin{tabular}{lcc}
\toprule
Method & Training & mAP (\%) \\
\midrule
ViTDeT-FT \citep{vitdet}           & $\checkmark$ & 2.4 \\
Detic \citep{detic}                & $\times$      & 0.0 \\
Detic-FT \citep{detic}             & $\checkmark$ & 3.8 \\
DE-ViT \citep{devit}               & $\times$      & 0.4 \\
DE-ViT-FT \citep{devit}            & $\checkmark$ & 0.6 \\
\midrule
Distill-cdfsod \citep{distill_cdfsod} & $\checkmark$ & nan \\
CD-ViTO \citep{cd_vito}            & $\checkmark$ & 3.6 \\
\midrule
\textbf{LAB-Det (ours)}          & $\times$      & \textbf{9.0} \\
\bottomrule
\end{tabular}
\label{tab:neudet}
\end{table}

\subsubsection{Experiments on NEU-DET}

On the NEU-DET dataset, which captures industrial surface defects with subtle textures and blurred boundaries, LAB-Det achieves 9.0 mAP in the one-shot setting, outperforming all baselines (\cref{tab:neudet}). Among ViT-based and open-vocabulary methods, the fine-tuned approach ViTDeT-FT \citep{vitdet} reaches only 2.4 mAP, while the training-free approach Detic \citep{detic} (0.0 mAP) and DE-ViT \citep{devit} (0.4 mAP) collapse due to pre-training mismatches with industrial imagery. Their fine-tuned counterparts (Detic-FT: 3.8 mAP, DE-ViT-FT: 0.6 mAP) improve slightly but remain far behind.  For CD-FSOD baselines, specialized cross-domain methods also fall short: CD-ViTO \citep{cd_vito} achieves 3.6 mAP, well below LAB-Det despite task-specific tuning.  

Overall, LAB-Det exceeds the strongest fine-tuned baseline (Detic-FT) by +5.2 mAP—more than doubling its performance. This gain stems from domain-aware prompts generated by DAM, which capture defect-specific cues (e.g., shape, texture) absent from raw class names. The PoE calibration further stabilizes tiny, low-contrast detections common in NEU-DET. Crucially, LAB-Det delivers these improvements without updating a single parameter, repeatedly reinforcing our central insight.

\begin{table}[t]
\centering
\caption{Ablation study of the overall training-free pipeline: comparisons between the Grounding DINO baseline (class-name prompts) and our full LAB-Det pipeline that incorporates all proposed design components. Training: $\checkmark$ represents fine-tuned, $\times$ represents training-free. The best results are highlighted in \textbf{bold}.}
\resizebox{\linewidth}{!}{
\begin{tabular}{l c c c c}
\toprule
\textbf{Method} & \textbf{Training} & \textbf{Prompt} & \textbf{mAP} & \textbf{AP$_{50}$} \\
\midrule
Grounding DINO & $\times$ & class name & 1.2 & 2.8 \\
\textbf{LAB-Det}  & $\times$ & support-to-text   & \textbf{9.0} & \textbf{17.8} \\
\bottomrule
\end{tabular}
}
\label{tab:neu_ablation_prompt}
\end{table}

\subsection{Ablation Study}

We assess the contribution of LAB-Det’s design by conducting an ablation study on NEU-DET. As a strong baseline, we evaluate Grounding DINO \citep{grounding_dino} with raw class names as prompts, leveraging its open-vocabulary capability without any adaptation. This baseline achieves only 1.2 mAP and 2.8 AP$_{50}$, highlighting that even a powerful generic model fails under severe domain shifts and blurred defect boundaries.  

By contrast, the full LAB-Det pipeline attains 9.0 mAP and 17.8 AP$_{50}$ (\cref{tab:neu_ablation_prompt}), yielding a +7.8 mAP gain. The improvement stems from projecting the one-shot exemplar into descriptive language, which injects domain-aware cues (e.g., shape, texture) that raw class names lack. These results validate that prompt-based linguistic adaptation can sharply reduce domain divergence, outperforming direct open-vocabulary detection in data-scarce specialized domains without parameter updates.

\begin{table}[t]
\centering
\caption{Ablation study on replacing raw class-name prompts with our support-driven text prompts. \textbf{LAB-Det-} represents a reduced version of our method that only incorporates the support-to-text prompt, while retaining other components from the Grounding DINO baseline. AR@1, AR@10, and AR@100 denote average recall when retaining at most 1, 10, or 100 predictions per image, respectively. Best results are highlighted in \textbf{bold}.}
\resizebox{\linewidth}{!}{
\begin{tabular}{l c c c c c}
\toprule
\textbf{Method} & \textbf{Prompt} & \textbf{AR@1} & \textbf{AR@10} & \textbf{AR@100} \\
\midrule
Grounding DINO & class name      & 6.3 & 23.5 & 38.5 \\
\textbf{LAB-Det-} & support-to-text & \textbf{13.5} & \textbf{30.4} & \textbf{40.9} \\
\bottomrule
\end{tabular}
}
\label{tab:neu_ar_ablation}
\end{table}

\paragraph{Effect of support-to-text prompts.}  
We isolate the impact of exemplar-to-language projection by comparing the Grounding DINO baseline (raw class-name prompts) with LAB-Det–, a reduced variant that replaces class names with support-derived text while keeping all other components identical. As shown in \cref{tab:neu_ar_ablation}, LAB-Det- significantly improves recall: AR@1 rises from 6.3 to 13.5 (+7.2), AR@10 from 23.5 to 30.4 (+6.9), and AR@100 from 38.5 to 40.9 (+2.4). These gains confirm that domain-aware descriptions inject richer cues—such as shape, texture, and color—that raw class names cannot capture, enhancing Grounding DINO’s phrase-sensitive matching. This ablation underscores that projecting exemplars into language reduces domain divergence and improves recall in high-shift settings without any training.

\begin{table}[t]
\centering
\caption{Ablation study on the proposed selective score calibration. “Selective Calibration’’ refers to the BLIP-based alignment applied only to small boxes. \textbf{AP\textsubscript{small}} is the average precision computed on small objects. Best results are highlighted in \textbf{bold}.}
\resizebox{\linewidth}{!}{
\begin{tabular}{l c c}
\toprule
\textbf{Method} & \textbf{Selective Calibration} & \textbf{AP$_{\text{small}}$} \\
\midrule
LAB-Det- (reduced variant) & $\times$ & 4.4  \\
LAB-Det\;\;(full pipeline)   & $\checkmark$ & \textbf{4.7} \\
\bottomrule
\end{tabular}
}
\label{tab:neu_ablation_calib}
\end{table}

\paragraph{Effect of selective score calibration.}  
We assess the impact of BLIP-based calibration by comparing LAB-Det– (without calibration) to the full LAB-Det pipeline on NEU-DET. As shown in \cref{tab:neu_ablation_calib}, calibration raises AP$_{\text{small}}$ from 4.4 to 4.7 (+0.3, $\sim$7\% relative). While modest in absolute terms, this improvement is consistent with our design goal: small proposals are particularly noisy, and fusing BLIP alignment provides a domain-sensitive correction via the product-of-experts rule. It is worth mentioning that calibration is applied only where needed, improving precision on vulnerable small objects without affecting larger ones or requiring any parameter updates.

\begin{figure}[t]
  \centering
   \includegraphics[width=1.0\linewidth]{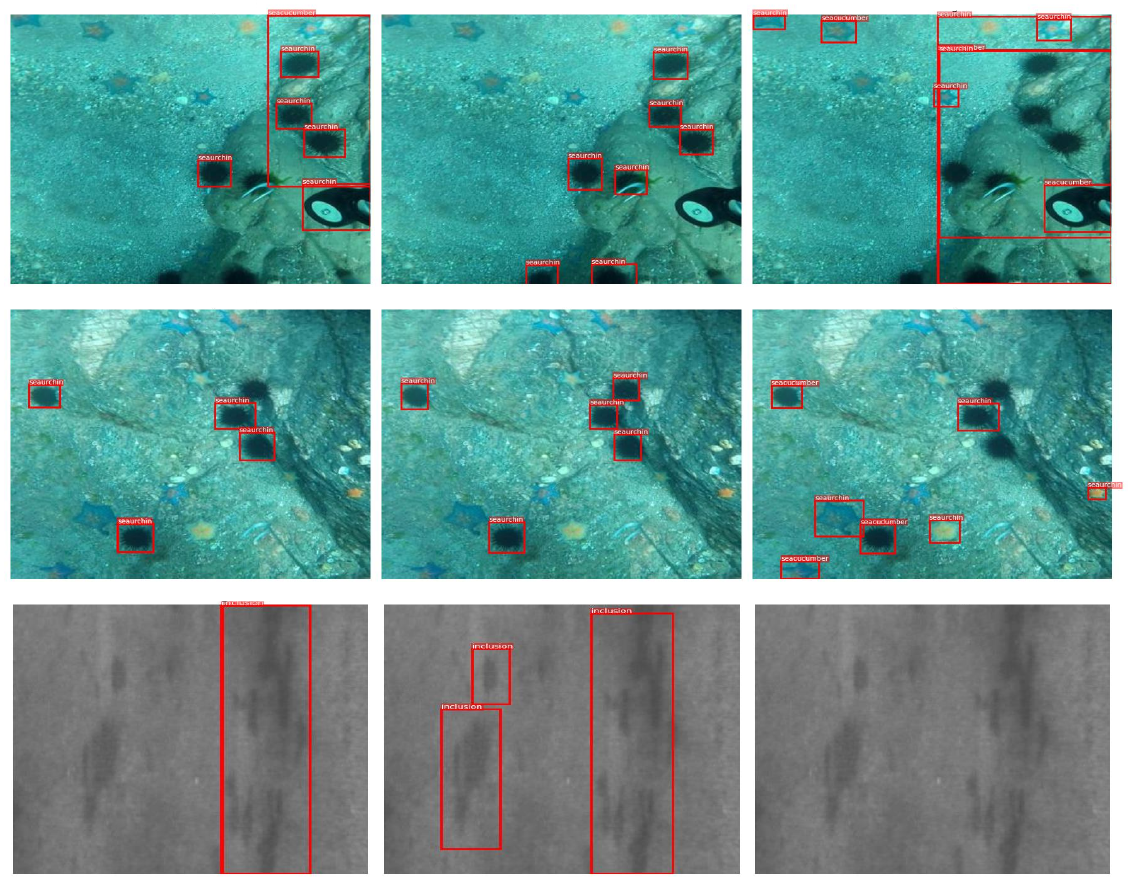}
   \caption{Qualitative comparison on UODD and NEU-DET datasets. From left to right: predictions from LAB-Det, ground-truth annotations, and Grounding DINO baseline, respectively.}
   \label{fig:display}
\end{figure}

\subsection{Visualization and Analysis}

To further illustrate LAB-Det’s behavior, we compare its predictions with ground truth (GT) and the Grounding DINO baseline in \cref{fig:display} (left to right: LAB-Det $\rightarrow$ GT $\rightarrow$ Grounding DINO). The first two rows show UODD examples, and the third row shows NEU-DET. For clarity, we omit trivial failure cases where both methods output image-sized boxes.  
Across the examples, LAB-Det produces tighter, more accurate boxes with fewer spurious detections than the class-name baseline, yielding visibly higher recall and precision. In underwater scenes, descriptive prompts highlight shape and texture cues that improve localization under color distortions. In industrial defects, calibration reduces false positives on tiny, low-contrast regions.  
These visualizations further strengthen our key hypothesis: language can act as a domain-invariant bridge, allowing frozen detectors to adapt training-free in high-IB domains, consistent with the quantitative gains reported in our main tables.  

\section{Conclusion and Discussion}

Foundation detectors pre-trained on general-domain data struggle in specialized environments such as underwater imagery and industrial defect inspection, where data is scarce and domain gaps are severe. To address this challenge, we introduced the task of \emph{training-free one-shot domain generalization for object detection} and proposed \textbf{LAB-Det}, a framework that leverages \emph{language as a domain-invariant bridge}.  
LAB-Det converts one-shot exemplars into descriptive prompts that guide a frozen detector through phrase-grounded proposals, transparent category assignment, and selective score calibration. Experiments on UODD and NEU-DET show that LAB-Det consistently surpasses strong fine-tuned CD-FSOD baselines and training-free foundation models, despite requiring no parameter updates.  
These results demonstrate that linguistic guidance can replace gradient-based adaptation, enabling efficient and interpretable generalization in high-IB domains. More broadly, LAB-Det highlights a paradigm shift: from weight updates toward prompt-based adaptation, expanding the practical utility of foundation models in specialized, data-scarce detection scenarios.

\textbf{Limitations and Future Work.} 
LAB-Det depends on an external captioner, where text quality might affect performance; integrating advanced vision-language models could yield richer prompts. Future work will extend to multiple-shot settings, video or 3D modalities, further advancing prompt-based frameworks for cross-domain perception.
{
    \small
    \bibliographystyle{ieeenat_fullname}
    \bibliography{main}
}


\end{document}